%% file: _Main.tex
\documentclass[11pt]{article}
\pdfoutput=1

\usepackage[preprint]{acl}

\usepackage{times}
\usepackage{latexsym}

\usepackage[T1]{fontenc}

\usepackage[utf8]{inputenc}

\usepackage{microtype}

\usepackage{inconsolata}

\usepackage{graphicx}
\usepackage{booktabs}
\usepackage{makecell}
\usepackage{array}
\usepackage{amsmath,amssymb}
\usepackage{multirow}
\usepackage{adjustbox}
\usepackage{xspace}
\usepackage{cleveref}
\usepackage{hyperref}  
\usepackage{float}
\usepackage{placeins} 
\usepackage{enumitem}
\usepackage[ruled,vlined,linesnumbered]{algorithm2e}
\usepackage[table]{xcolor}

\newcommand{\predictorN}{FRAG\xspace}
\newcommand{\pruningN}{FRP\xspace}

\definecolor{fst}{rgb}{0.909, 0.504, 0.588}
\definecolor{sed}{rgb}{0.994, 0.806, 0.742}
\definecolor{thd}{rgb}{0.999, 0.966, 0.921}

\title{Distance Is Not Enough: Forget-Retain Alignment Gap \\ Predicts LLM Relearning Robustness}

\author{
  \textbf{Yi Chen\textsuperscript{1*}},
  \textbf{Hanna Hsieh\textsuperscript{1*}},
  \textbf{Shuhong Liu\textsuperscript{2}},
  \textbf{Chuanbo Hua\textsuperscript{1}},
\\
  \textbf{Zihan Ma\textsuperscript{1}},
  \textbf{Kun Wang\textsuperscript{1}},
  \textbf{Joo-Young Kim\textsuperscript{1}}
\\
\\
  \textsuperscript{1}KAIST \quad
  \textsuperscript{2}The University of Tokyo
\\
  \small{
    \texttt{\{chenyi, hihahanaisme, cbhua, zihanma, walkerwang,
      jooyoung1203\}@kaist.ac.kr}
  }
\\
  \small{
    \texttt{s-liu@mi.t.u-tokyo.ac.jp}
  }
}

\begin{document}
\maketitle
\renewcommand{\thefootnote}{\fnsymbol{footnote}}
\footnotetext[1]{Equal contribution.}
\renewcommand{\thefootnote}{\arabic{footnote}}
\setcounter{footnote}{0}
\input{latex/0_Abstract}

\input{latex/1_Introduction}
\input{latex/2_Related}

\input{latex/3_Method}
\input{latex/4_Experiment}

\input{latex/5_Ablation}
\input{latex/6_Conclusion}
\input{latex/7_Limitation}

\section*{Acknowledgments}
This work was supported by the Institute of Information \& Communications
Technology Planning \& Evaluation (IITP) grants funded by the Korea government
(MSIT) (\mbox{IITP-2026-RS-2020-II201847}, Information Technology Research
Center; \mbox{IITP-2026-RS-2023-00256472}, Graduate School of Artificial
Intelligence Semiconductor).

\bibliography{custom}

\appendix
\input{latex/99_Appendix}

\end{document}

%% file: latex/0_Abstract.tex
\begin{abstract}

Machine unlearning aims to make a model forget specific data, yet unlearned LLMs often fail to stay unlearned: brief fine-tuning can revive removed knowledge. Existing robustness predictors rely on global weight-space displacement, but distance alone can be misleading when random or destructive updates collapse performance. We argue that relearning robustness depends on update structure: robust unlearning should affect forget-critical weights while sparing retain-critical ones. We introduce the Forget-Retain Alignment Gap (\predictorN), a training-free predictor that scores an update's forget-retain alignment without running a relearning attack, and separates selective from dense updates more reliably than global distance. Building on the forget-critical, retain-sparing principle, Forget-Retain Pruning (\pruningN) improves relearning robustness. Our results suggest that weight selectivity better explains robustness than distance alone. Our code is available at \url{https://github.com/Yi1-Chen/FRAG}.

\end{abstract}

%% file: latex/1_Introduction.tex
\section{Introduction}

\looseness=-1 Machine unlearning aims to remove the influence of specified data from a trained model while preserving its behavior on the remaining retain data \citep{maini2024tofu,li2024wmdp}. However, for large language models (LLMs), forgetting at edit time is often fragile: even without access to the forgotten examples, subsequent fine-tuning on benign retain data can revive the supposedly removed knowledge, forming a relearning attack \citep{hu2024jogging,lynch2024eight}. This relearning behavior exposes a central weakness of current unlearning methods: successful forgetting at edit time does not necessarily imply robustness against relearning attacks \citep{lucki2025an,che2025model,deeb2024unlearning}.

Recent work argues that relearning robustness can be improved by moving the unlearned model farther from the original model in weight space, making the removed knowledge harder to recover under relearning attacks \citep{siddiqui2025dormant}. This motivates using the global $\ell_2$ weight-space distance between the original and unlearned weights as a simple robustness predictor. However, distance alone can be misleading: it measures how far the weights move, but not which weights move. For example, a random or destructive update can yield a large $\ell_2$ displacement and appear robust to a distance-based predictor, while collapsing retain or forget performance and producing a model that is far from the original but not meaningfully unlearned. Robust unlearning should depend not only on displacement magnitude, but on whether the update concentrates on forget-critical weights while sparing retain-critical ones.

This observation motivates a proxy that diagnoses where the unlearning update is concentrated, not only how large it is.
We introduce the Forget--Retain Alignment Gap (\predictorN), a training-free scalar proxy for predicting relearning robustness that measures whether the update aligns more with forget-critical than retain-critical weights.
By penalizing retain-side disruption, \predictorN avoids rewarding collapsed models and better reflects practical unlearning robustness.

Across diverse unlearning methods \citep{zhang2024npo,maini2024tofu,li2024wmdp,pochinkov2024selective,jang-etal-2023-knowledge}, benchmarks, and model families, we show that \predictorN correlates with empirical relearning robustness more reliably than global distance-based predictors.
To show that this principle is actionable rather than merely diagnostic, we instantiate it as Forget-Retain Pruning (\pruningN), which selectively targets forget-critical weights while avoiding retain-critical ones.
\pruningN improves robustness under relearning attacks, tracing a robustness--utility frontier that dominates strong baselines at every matched utility level, suggesting that which weights move matters more than distance alone.

Our contributions are summarized as follows:
\begin{itemize}[itemsep=0.15em, topsep=0.2em, parsep=0pt, partopsep=0pt]
    \item We revisit relearning robustness from a weight-selectivity perspective, showing that global distance alone cannot distinguish selective unlearning updates from random or retain-damaging perturbations.
    \item \looseness=-1 We introduce \predictorN, a training-free proxy that predicts relearning robustness by diagnosing whether an update is forget-critical and retain-sparing.
    \item As an application of the same principle, we propose \pruningN, which improves relearning robustness at a controllable utility cost.
\end{itemize}

%% file: latex/2_Related.tex
\section{Related Work}
\label{sec:related}

\paragraph{Machine Unlearning for LLMs.}
\label{sec:related-methods}
LLM unlearning removes a forget set's influence while preserving
retain utility, judged jointly on the two~\citep{maini2024tofu}.
Most methods optimize a forget-derived loss: likelihood suppression
(GA, \citealp{jang-etal-2023-knowledge}; GradDiff,
\citealp{liu2022continual}), preference optimization
(NPO, \citealp{zhang2024npo}; SimNPO, \citealp{fan2025simnpo}), and
representation engineering (RMU; \citealp{li2024wmdp}). Others edit forget-related weights directly (pruning, attribution)
or drop the retain set~\citep{wang2024flat} or act only at
inference~\citep{pawelczyk2024incontext}.

\paragraph{Relearning robustness.}
\label{sec:related-attacks}
Unlearned LLMs recover forgotten knowledge under modest extra
training~\citep{hu2024jogging,lynch2024eight,lucki2025an,
schwinn2024soft,patil2024can,deeb2024unlearning,che2025model},
suggesting edit-time forgetting suppresses rather than removes
knowledge. Proposed defenses include sharpness-aware
unlearning~\citep{fan2025towards}, latent adversarial
training~\citep{sheshadri2025lat}, tamper-resistant
safeguards~\citep{tamirisa2025tar}, and localized
edits~\citep{guo2025mechanistic}. Closest to us,
\citet{siddiqui2025dormant} tie robustness to weight-space
displacement; we show a scalar global $\ell_2$ distance is insufficient: which
weights move, not how far, governs robustness.

\input{Figures/intro}

\paragraph{Weight Importance and Pruning.}
\label{sec:related-pruning}
\looseness=-1 Unstructured LLM pruning scores weights by activations
(Wanda; \citealp{sun2024wanda}), relative importance
(RIA; \citealp{zhang2024plugandplay}), or second-order reconstruction
(SparseGPT; \citealp{frantar2023sparsegpt}). This importance also
varies with the input, as online structured pruning adapts channel
masks to the current context (POP; \citealp{chen2026pop}). A parallel line localizes
knowledge to FFN memories \citep{geva-etal-2021-transformer}, neurons
\citep{dai-etal-2022-knowledge}, and MLP modules
\citep{meng2022locating}. For unlearning, Selective
Pruning~\citep{pochinkov2024selective} uses forget--retain activation
contrast, SSD~\citep{foster2024ssd} dampens forget weights via Fisher
information (vision), SalUn~\citep{fan2024salun} uses gradient-based
weight saliency, and WAGLE~\citep{jia2024wagle} uses
gradient attribution; \citet{jia2023sparsity} show in vision that
sparsity alone eases unlearning. None target the forget--retain
alignment structure governing relearning robustness; \pruningN builds on
the importance contrast of Selective Pruning and applies it to relearning
robustness (\S\ref{sec:frp}).

%% file: Figures/intro.tex
\begin{figure*}[t]
  \centering
  \includegraphics[width=\textwidth]{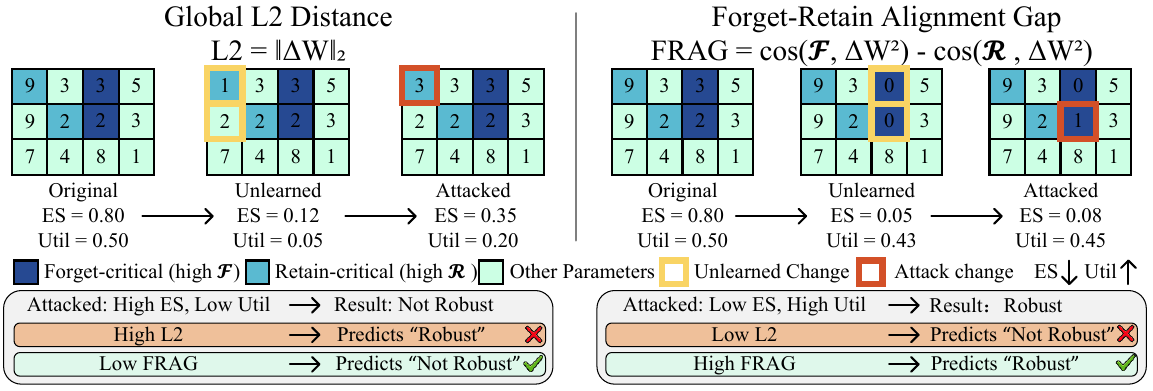}
    \vspace{-20pt}
\caption{
Illustration of global $L_2$ and \predictorN\ as attack-free predictors of relearning robustness.
Large $L_2$ can falsely suggest robustness when edits hit retain-critical weights, while small $L_2$ can miss robust forget-critical edits.
\predictorN\ captures both cases by measuring whether updates target forget-critical while sparing retain-critical weights.
}
  \label{fig:Intro}
  \vspace{-10pt}
\end{figure*}

%% file: latex/3_Method.tex
\section{Method}
\label{sec:method}

We develop a weight-selective view of relearning robustness.
Our starting point is that a robust unlearned model should not merely move far from the original model; its update should be concentrated on forget-critical weights while avoiding retain-critical ones.
Based on this property, we first define an attack-free robustness prediction problem, then introduce \predictorN as a diagnostic proxy.
Finally, we instantiate the same principle as \pruningN, which directly constructs more robust unlearned models by enforcing this selective update structure.

\subsection{Problem Formulation}
\label{sec:problem}

\looseness=-1 Let $M_0$ and $M_u$ denote the original and unlearned models with parameters $\theta_0$ and $\theta_u$.
Given a forget set $\mathcal{D}_f$ and retain set $\mathcal{D}_r$, unlearning aims to remove the influence of $\mathcal{D}_f$ while maintaining retain-side behavior on $\mathcal{D}_r$.
After unlearning, $M_u$ may face a relearning attack by fine-tuning on an attack set $\mathcal{D}_a$ drawn from $\mathcal{D}_r$, $\mathcal{D}_f$, or their mixture, producing an attacked model $M_a$.

A robust unlearned model should resist recovery of forgotten knowledge while maintaining retain-side utility.
Thus, robustness is not captured by post-attack forgetting alone; a utility-collapsed model is not meaningfully robust.
Our goal is to identify attack-free weight-space properties that predict such robustness.
Given $M_0$, $M_u$, and small calibration sets
$\mathcal{D}_f^{\mathrm{cal}},\mathcal{D}_r^{\mathrm{cal}}$, we seek an attack-free scoring function
{\setlength{\abovedisplayskip}{3pt}
\setlength{\belowdisplayskip}{3pt}
\begin{equation}
\label{eq:robustness_proxy}
    \phi:
    (M_0,M_u,\mathcal{D}_f^{\mathrm{cal}},\mathcal{D}_r^{\mathrm{cal}})
    \mapsto \mathbb{R},
\end{equation}}
where $\mathbb{R}$ denotes the real numbers and higher scores indicate stronger predicted robustness under relearning attacks. Unlike global $\ell_2$ distance $\|\theta_u-\theta_0\|_2$, which measures only update magnitude, we focus on where the update is concentrated.
\subsection{Predicting Robustness: \predictorN}
\label{sec:frag}

A weight is \emph{forget-critical} if its magnitude and input-channel activation indicate greater importance on forget than on retain data; \emph{retain-critical} is defined symmetrically. Both denote \emph{relative}, data-dependent importance rather than weights exclusive to one set: nearly every weight carries some of both, and what matters is the ratio.

For layer $\ell$, let $W_0^\ell$ and $W_u^\ell$ be the original and unlearned weights. The unlearning update is
\begingroup
\setlength{\abovedisplayskip}{3pt}
\setlength{\belowdisplayskip}{3pt}
\begin{equation}
\label{eq:update}
    \Delta W^\ell = W_u^\ell - W_0^\ell .
\end{equation}
\endgroup
To assess whether the update is robustly structured, we compare $(\Delta W^\ell)^2$ with forget- and retain-critical weight importance.

For each input channel $j$, we collect activation norms on forget and retain calibration data, denoted $x^{f,\ell}_j$ and $x^{r,\ell}_j$. Following the weight importance~\citep{sun2024wanda}, we define
\begingroup
\setlength{\abovedisplayskip}{3pt}
\setlength{\belowdisplayskip}{3pt}
\begin{align}
\label{eq:forget_importance}
    \mathcal{F}_{ij}^\ell
    &= |(W_0^\ell)_{ij}|\,
    \frac{x^{f,\ell}_j}{x^{r,\ell}_j+\epsilon}, \\[-0.3em]
\label{eq:retain_importance}
    \mathcal{R}_{ij}^\ell
    &= |(W_0^\ell)_{ij}|\,
    \frac{x^{r,\ell}_j}{x^{f,\ell}_j+\epsilon}.
\end{align}
\endgroup
Here, $\mathcal{F}^\ell$ and $\mathcal{R}^\ell$ denote forget- and retain-critical weight importance, respectively.
Let $D=(\Delta W)^2$ denote the squared update after aggregating selected layers.
We use cosine similarity because it is scale-invariant, measuring alignment rather than update magnitude:
\begingroup
\setlength{\abovedisplayskip}{3pt}
\setlength{\belowdisplayskip}{3pt}
\begin{equation}
\label{eq:frag}
\begin{aligned}
A_f &= \cos(\mathcal{F},D),\\[-0.15em]
A_r &= \cos(\mathcal{R},D),\\[-0.15em]
\mathrm{FRAG} &= A_f-\gamma A_r .
\end{aligned}
\end{equation}
\endgroup
Here, $A_f$ and $A_r$ are forget- and retain-update alignment scores.
A high \predictorN indicates the update aligns with forget-critical
weights while avoiding retain-critical ones. The retain term ($\gamma\!=\!1$;
App.~\ref{app:fria_recipe}) prevents forget-only alignment from
rewarding destructive updates that damage utility.

Fine-tuning can only move a weight that the fine-tuning data actually uses: for a linear layer, the gradient of $W_{ij}$ carries a factor $x_j$, the activation of its input channel. Weights that fire on forgotten content but not on retain content are therefore inert under a retain-only relearning attack, and an edit placed there survives it. \predictorN scores exactly this placement, which is why an attack-free score can anticipate the attack. The argument is local and first-order, and it weakens once the attacker also holds forget data, which reactivates those channels; we therefore evaluate \predictorN\ under that stronger attack (Table~\ref{tab:rank_corr}).

\input{Tables/Attack_Main_Table}

\input{Tables/t2_test}

\input{Figures/FRIP_Algo}
\looseness=-1 Computing \predictorN only requires calibration forward passes and a weight comparison between $M_0$ and $M_u$; it needs no relearning attack or additional optimization. Unless specified otherwise, we score attention and MLP projection layers. Because \predictorN measures directional alignment, diffuse dense updates align weakly and receive substantially smaller scores; it therefore separates selective from dense updates reliably, while resolving differences \emph{among} dense methods only coarsely. See Appendix~\ref{app:fria_recipe} for details.

\subsection{Achieving Robustness: \pruningN}
\label{sec:frp}

The same weight-selective principle can be used to construct robust unlearned models.
We propose \pruningN, which prunes weights that are forget-important, retain-unimportant, and large enough to induce a meaningful edit.

For each target module, \pruningN computes the weight-aware importance scores $\mathcal{F}$ and $\mathcal{R}$ from Eq.~\eqref{eq:forget_importance}--\eqref{eq:retain_importance}.
For each output row $i$, it scores each weight $W_{ij}$ by
\begingroup
\setlength{\abovedisplayskip}{3pt}
\setlength{\belowdisplayskip}{3pt}
\begin{equation}
\label{eq:frp_score}
\begin{aligned}
S_{ij}
=&\ \operatorname{rank}_j(\mathcal{F}_{ij})
-\beta\,\operatorname{rank}_j(\mathcal{R}_{ij}) \\
&+\lambda\,\operatorname{rank}_j(|W_{ij}|).
\end{aligned}
\end{equation}
\endgroup
Here, $\operatorname{rank}_j(\cdot)$ ranks entries within the same output row, with larger values receiving larger ranks.
The three terms favor forget-critical weights, penalize retain-critical weights, and add a magnitude prior so that pruning produces a nontrivial edit. The hyperparameters $\beta$ and $\lambda$ control the retain penalty and magnitude prior, respectively.
\looseness=-1 This weight-level, rank-space scoring distinguishes \pruningN from Selective Pruning~\citep{pochinkov2024selective}, which thresholds a raw importance ratio at the neuron level and is not evaluated for relearning robustness.
Finally, \pruningN prunes the top $\lfloor\rho d_i\rfloor$ weights in each row according to $S_{i,:}$, as shown in Algorithm~\ref{alg:frip}. See Appendix~\ref{app:frip} for details and ablation study.

%% file: Tables/Attack_Main_Table.tex
\begin{table*}[!htbp]
\centering
\footnotesize
\setlength{\tabcolsep}{2.7pt}
\renewcommand{\arraystretch}{1.0}
\begin{adjustbox}{max width=\textwidth,center}
\begin{tabular}{@{}l|cc|ccc|ccc|ccc|ccc|cc@{}}
\toprule
\multirow{2}{*}{\textbf{Method}}
 & \multicolumn{2}{c|}{\textbf{Unlearned}}
 & \multicolumn{3}{c|}{\textbf{\makecell{Retain Attack}}}
 & \multicolumn{3}{c|}{\textbf{\makecell{Forget Attack}}}
 & \multicolumn{3}{c|}{\textbf{\makecell{Forget+Retain Attack}}}
 & \multicolumn{3}{c|}{\textbf{Average}}
 & \multicolumn{2}{c}{\textbf{Predictor}} \\
 & ES\,$\downarrow$ & Util\,$\uparrow$
 & ES\,$\downarrow$ & $\Delta$ES\,$\downarrow$ & Util\,$\uparrow$
 & ES\,$\downarrow$ & $\Delta$ES\,$\downarrow$ & Util\,$\uparrow$
 & ES\,$\downarrow$ & $\Delta$ES\,$\downarrow$ & Util\,$\uparrow$
 & $\overline{\text{ES}}\!\downarrow$ & $\overline{\Delta\text{ES}}\!\downarrow$ & $\overline{\text{Util}}\!\uparrow$
 & L2\,$\uparrow$ & \predictorN$\uparrow$ \\
\midrule
\multicolumn{17}{c}{\textit{LLaMA-3.2-1B}} \\
\midrule
Retain   & 0.064 & 0.596 & 0.063 & -0.001 & 0.597 & 0.081 & 0.017 & 0.580 & 0.071 & 0.007 & 0.597 & 0.072 & 0.008 & 0.591 & - & - \\
\midrule
GA       & \colorbox{sed}{0.086} & 0.199 & 0.284 & 0.198 & \colorbox{thd}{0.598} & \colorbox{thd}{0.153} & \colorbox{thd}{0.067} & 0.359 & 0.401 & 0.315 & \colorbox{thd}{0.598} & 0.279 & 0.193 & 0.518 & 0.875 & 0.002 \\
GradDiff & 0.123 & \colorbox{sed}{0.498} & 0.256 & 0.134 & \colorbox{fst}{0.602} & 0.264 & 0.142 & \colorbox{fst}{0.591} & 0.441 & 0.318 & \colorbox{fst}{0.600} & 0.320 & 0.198 & \colorbox{sed}{0.598} & 0.569 & 0.003 \\
NPO      & 0.126 & 0.487 & 0.215 & 0.089 & \colorbox{sed}{0.603} & 0.197 & 0.071 & \colorbox{thd}{0.547} & 0.331 & 0.205 & \colorbox{sed}{0.600} & 0.248 & 0.122 & \colorbox{thd}{0.583} & 0.748 & 0.000 \\
RMU      & 0.105 & \colorbox{fst}{0.561} & 0.412 & 0.307 & \colorbox{fst}{0.602} & 0.420 & 0.315 & \colorbox{sed}{0.579} & 0.742 & 0.637 & \colorbox{sed}{0.601} & 0.525 & 0.420 & \colorbox{fst}{0.594} & 0.769 & 0.004 \\
SP       & 0.124 & 0.486 & \colorbox{thd}{0.171} & \colorbox{thd}{0.047} & 0.521 & 0.200 & 0.076 & 0.497 & \colorbox{thd}{0.237} & \colorbox{thd}{0.113} & 0.520 & \colorbox{thd}{0.203} & \colorbox{thd}{0.079} & 0.513 & \colorbox{fst}{120.6} & \colorbox{thd}{0.393} \\
\midrule
\textbf{\pruningN}~($\beta{=}0.00$)
         & \colorbox{fst}{0.055} & 0.384 & \colorbox{fst}{0.079} & \colorbox{fst}{0.024} & 0.464 & \colorbox{fst}{0.071} & \colorbox{fst}{0.015} & 0.407 & \colorbox{fst}{0.104} & \colorbox{fst}{0.049} & 0.463 & \colorbox{fst}{0.085} & \colorbox{fst}{0.029} & 0.444 & \colorbox{sed}{111.5} & \colorbox{fst}{3.159} \\
\textbf{\pruningN}~($\beta{=}0.15$)
         & \colorbox{thd}{0.099} & \colorbox{thd}{0.494} & \colorbox{sed}{0.133} & \colorbox{sed}{0.034} & 0.526 & \colorbox{sed}{0.138} & \colorbox{sed}{0.039} & 0.498 & \colorbox{sed}{0.201} & \colorbox{sed}{0.102} & 0.526 & \colorbox{sed}{0.157} & \colorbox{sed}{0.058} & 0.517 & \colorbox{thd}{81.4} & \colorbox{sed}{3.020} \\
\midrule
\midrule
\multicolumn{17}{c}{\textit{LLaMA-3.2-3B}} \\
\midrule
Retain   & 0.064 & 0.658 & 0.071 & 0.007 & 0.655 & 0.085 & 0.022 & 0.639 & 0.082 & 0.019 & 0.651 & 0.080 & 0.016 & 0.649 & - & - \\
\midrule
GA       & 0.120 & 0.384 & 0.331 & 0.211 & \colorbox{fst}{0.671} & 0.222 & 0.102 & 0.548 & 0.497 & 0.376 & \colorbox{fst}{0.673} & 0.350 & 0.230 & 0.630 & 1.363 & 0.002 \\
GradDiff & 0.195 & 0.584 & 0.380 & 0.184 & \colorbox{thd}{0.658} & 0.454 & 0.259 & \colorbox{thd}{0.654} & 0.553 & 0.358 & \colorbox{thd}{0.660} & 0.462 & 0.267 & \colorbox{thd}{0.657} & 0.937 & 0.005 \\
NPO      & \colorbox{thd}{0.082} & \colorbox{sed}{0.663} & \colorbox{sed}{0.108} & \colorbox{fst}{0.026} & \colorbox{sed}{0.665} & 0.175 & \colorbox{thd}{0.093} & \colorbox{sed}{0.649} & \colorbox{thd}{0.281} & \colorbox{thd}{0.198} & \colorbox{sed}{0.664} & \colorbox{thd}{0.188} & \colorbox{thd}{0.106} & \colorbox{sed}{0.660} & 1.506 & 0.002 \\
RMU      & \colorbox{fst}{0.054} & \colorbox{fst}{0.664} & \colorbox{thd}{0.123} & 0.068 & \colorbox{sed}{0.664} & \colorbox{thd}{0.139} & \colorbox{sed}{0.085} & \colorbox{fst}{0.663} & 0.755 & 0.700 & \colorbox{sed}{0.664} & 0.339 & 0.284 & \colorbox{fst}{0.664} & 1.488 & 0.007 \\
SP       & 0.187 & 0.589 & 0.299 & 0.112 & 0.613 & 0.431 & 0.244 & 0.597 & 0.443 & 0.257 & 0.618 & 0.391 & 0.204 & 0.610 & \colorbox{fst}{191.9} & \colorbox{thd}{0.941} \\
\midrule
\textbf{\pruningN}~($\beta{=}0.00$)
         & \colorbox{sed}{0.068} & 0.499 & \colorbox{fst}{0.096} & \colorbox{sed}{0.028} & 0.569 & \colorbox{fst}{0.107} & \colorbox{fst}{0.039} & 0.522 & \colorbox{fst}{0.147} & \colorbox{fst}{0.079} & 0.570 & \colorbox{fst}{0.117} & \colorbox{fst}{0.049} & 0.553 & \colorbox{sed}{182.2} & \colorbox{sed}{3.538} \\
\textbf{\pruningN}~($\beta{=}0.05$)
         & 0.093 & 0.555 & 0.127 & \colorbox{thd}{0.034} & 0.598 & \colorbox{sed}{0.132} & \colorbox{fst}{0.039} & 0.561 & \colorbox{sed}{0.195} & \colorbox{sed}{0.102} & 0.600 & \colorbox{sed}{0.151} & \colorbox{sed}{0.058} & 0.587 & \colorbox{thd}{162.3} & \colorbox{fst}{3.738} \\
\bottomrule
\end{tabular}
\end{adjustbox}
 \vspace{-5pt}
\caption{
Relearning robustness on TOFU averaged over forget-set sizes.
We report post-attack ES, $\Delta$ES, and utility under retain, forget, and forget+retain attacks, along with attack-free predictors.
\pruningN achieves the lowest post-attack ES/$\Delta$ES with favorable robustness-utility tradeoff. \predictorN better identifies robust updates than global $\ell_2$ distance.
}
\label{tab:tofu_attacks_avg_byMethod}
\vspace{-0.1in}
\end{table*}

%% file: Tables/t2_test.tex
\begin{table}[!htbp]
\centering
\footnotesize
\setlength{\tabcolsep}{2pt}
\renewcommand{\arraystretch}{1.05}
\begin{adjustbox}{max width=\columnwidth,center}
\begin{tabular}{@{}l|cc|ccc|cc@{}}
\toprule
\multirow{2}{*}{\textbf{Method}}
 & \multicolumn{2}{c|}{\textbf{Unlearned}}
 & \multicolumn{3}{c|}{\textbf{Retain Attack}}
 & \multicolumn{2}{c}{\textbf{Predictor}} \\
 & Acc\,$\downarrow$ & MMLU\,$\uparrow$
 & Acc\,$\downarrow$ & $\Delta$Acc\,$\downarrow$ & MMLU\,$\uparrow$
 & L2\,$\uparrow$ & \predictorN\,$\uparrow$ \\
\midrule
Ref & 0.583 & 0.788 & 0.581 & -0.002 & 0.792 & --- & --- \\
\midrule
RMU
& \colorbox{sed}{0.479} & \colorbox{fst}{0.782}
& \colorbox{thd}{0.552} & \colorbox{thd}{+0.073} & \colorbox{fst}{0.791}
& \colorbox{thd}{26.3} & \colorbox{thd}{0.185} \\
SP
& \colorbox{thd}{0.515} & \colorbox{sed}{0.764}
& \colorbox{sed}{0.513} & \colorbox{sed}{-0.002} & \colorbox{sed}{0.771}
& \colorbox{fst}{464.1} & \colorbox{sed}{3.149} \\
\midrule
\textbf{\pruningN}~($\beta{=}0$)
& \colorbox{fst}{0.429} & \colorbox{thd}{0.668}
& \colorbox{fst}{0.417} & \colorbox{fst}{-0.012} & \colorbox{thd}{0.707}
& \colorbox{sed}{443.4} & \colorbox{fst}{9.198} \\
\bottomrule
\end{tabular}
\end{adjustbox}
\caption{
Cross-family validation on WMDP-cyber with Qwen2.5-14B-Instruct.
}
\label{tab:wmdp_relearn_tamper}
\vspace{-0.2in}
\end{table}

%% file: Figures/FRIP_Algo.tex
\begin{algorithm}[t]
\caption{Forget--Retain Pruning (\pruningN)}
\label{alg:frip}
\scriptsize
\SetAlgoNlRelativeSize{-1}
\SetNlSkip{0.35em}
\SetInd{0.35em}{0.55em}
\KwIn{$\theta_0$, $\mathcal{D}_f,\mathcal{D}_r$, modules $\mathcal{M}$, sparsity $\rho$, retain penalty $\beta$, magnitude weight $\lambda$}
\KwOut{$\theta_u$}
\For{$m\in\mathcal{M}$ with $W\in\mathbb{R}^{d_o\times d_i}$}{
    $x^f,x^r \leftarrow$ input-channel norms on $\mathcal{D}_f,\mathcal{D}_r$\;
    $\mathcal{F}_{ij}\leftarrow |W_{ij}|x^f_j/(x^r_j+\epsilon)$;\quad
    $\mathcal{R}_{ij}\leftarrow |W_{ij}|x^r_j/(x^f_j+\epsilon)$\;
    \For{$i=1,\ldots,d_o$}{
        $S_{ij}\leftarrow
        \operatorname{rank}_j(\mathcal{F}_{ij})
        -\beta\operatorname{rank}_j(\mathcal{R}_{ij})
        +\lambda\operatorname{rank}_j(|W_{ij}|)$\;
        $\mathcal{P}_i\leftarrow\operatorname{TopK}(S_{i,:},\lfloor\rho d_i\rfloor)$\;
        $W_{i,\mathcal{P}_i}\leftarrow 0$\;
    }
}
\Return $\theta_u$\;
\end{algorithm}

%% file: latex/4_Experiment.tex
\section{Experiments}
\label{sec:experiments}

\paragraph{Evaluation Setups.}
All experiments use OpenUnlearning~\cite{dorna2026openunlearning}. 
We evaluate TOFU~\cite{maini2024tofu} on LLaMA-3.2-1B/3B~\cite{grattafiori2024llama3} across \texttt{forget01}/\texttt{05}/\texttt{10}, using retain, forget, and forget+retain relearning attacks, and WMDP-cyber~\cite{li2024wmdp} on Qwen2.5-14B-Instruct~\cite{qwen2024qwen25}; Appendix~\ref{app:muse} adds MUSE-News~\cite{shi2025muse}.
Baselines include GA~\citep{jang-etal-2023-knowledge}, GradDiff~\citep{liu2022continual}, NPO~\citep{zhang2024npo}, RMU~\citep{li2024wmdp}, and SP~\citep{pochinkov2024selective}. 
We report ES/$\Delta$ES/utility on TOFU and Acc/$\Delta$Acc/MMLU on WMDP, and compare global $\ell_2$ distance~\cite{siddiqui2025dormant} with \predictorN as robustness predictors. 
Best results are shaded \colorbox{fst}{first}, \colorbox{sed}{second}, \colorbox{thd}{third}; details are in Appendix~\ref{app:frip}.

\input{Figures/ranking}

\paragraph{Main Results.}
\label{sec:main_results}
Table~\ref{tab:tofu_attacks_avg_byMethod} shows that \pruningN consistently achieves the best average post-attack ES and $\Delta$ES on TOFU across both model sizes and three relearning attacks. 
Although SP obtains very large global $\ell_2$ distance, its performance is worse than \pruningN, showing that distance alone misranks update quality. 
\looseness=-1 Table~\ref{tab:wmdp_relearn_tamper} confirms the same trend on WMDP-cyber: \pruningN achieves the lowest cyber accuracy after retain-set relearning, and \predictorN assigns it the highest robustness score despite SP having larger $\ell_2$ distance. This comes at a cost: \pruningN's MMLU drop exceeds that of RMU and SP, so \pruningN traces a robustness--utility frontier rather than dominating on both axes. See Appendix~\ref{app:ablation} for guidance on choosing $\beta$ and $\rho$, and Appendix~\ref{app:extended} for more results.
\paragraph{Predictor Analysis.}
\label{sec:predictor_analysis}
Figure~\ref{fig:ranking} isolates predictor behavior on TOFU forget10 checkpoints with noise-based collapsed controls.
After per-predictor normalization, global $\ell_2$ and individual cosine terms peak on collapsed perturbations, while \predictorN peaks on the low-ES, non-collapsed \pruningN checkpoint.
This shows why retain-aware alignment is necessary.
\input{Tables/rank_corr}
Table~\ref{tab:rank_corr} makes the comparison quantitative over healthy checkpoints only, with collapsed and noise controls removed. \predictorN reaches $\rho=-0.78$ pooled against $-0.36$ for global $\ell_2$, and the gap widens at 3B, where $\ell_2$ falls to $-0.17$. Dropping every \pruningN checkpoint leaves \predictorN at $-0.74$ while $\ell_2$ falls to $-0.10$, reversing sign at 3B, so the ranking power does not come from \predictorN scoring the method built on it. See more in Appendix~\ref{app:noise_ablation}.

%% file: Figures/ranking.tex
\begin{figure}[t]
    \centering
    \includegraphics[width=\linewidth]{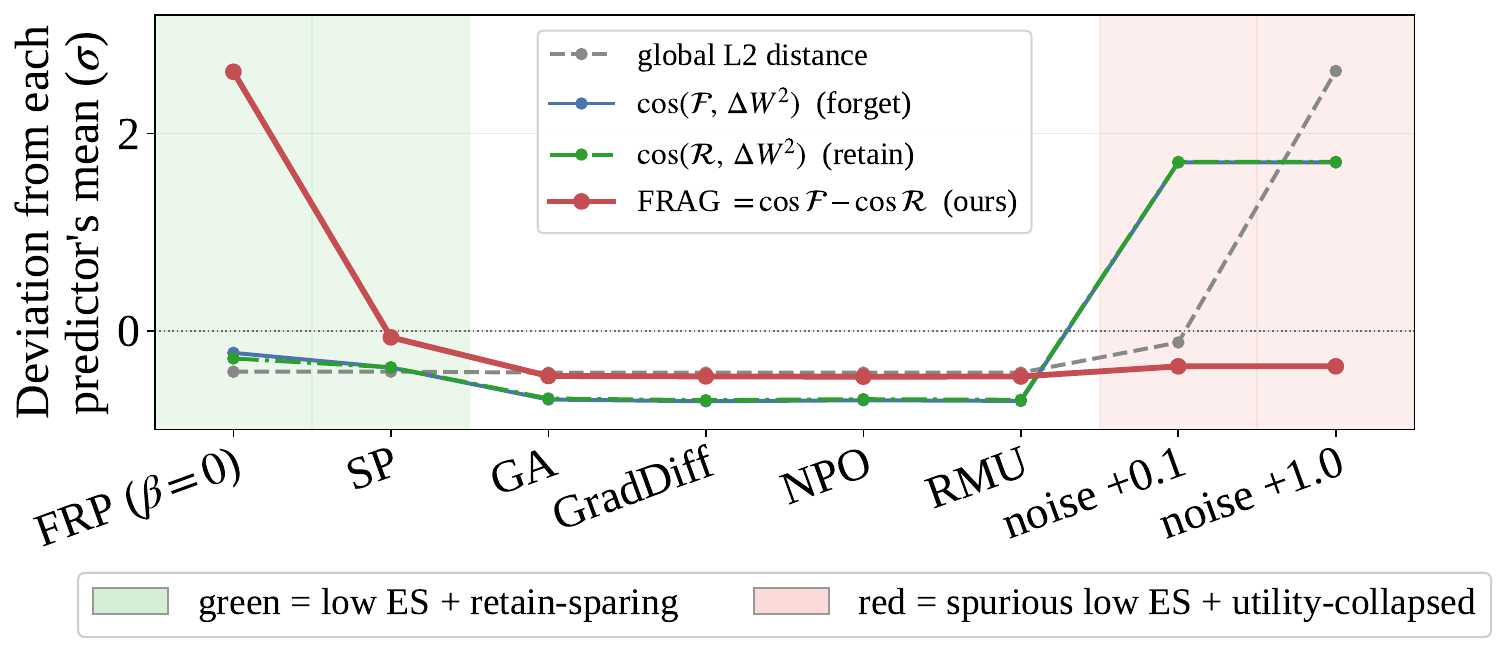}
    \vspace{-0.3in}
\caption{
Curves are normalized by their own mean/std across unlearned checkpoints and noise controls.
Global $\ell_2$ and individual cosine terms peak on utility-collapsed perturbations, while \predictorN\ peaks on the low-ES, retain-sparing \pruningN checkpoint.
}
    \label{fig:ranking}
\end{figure}

%% file: Tables/rank_corr.tex
\begin{table}[t]
\centering
\small
\setlength{\tabcolsep}{5pt}
\begin{tabular}{lcccc}
\toprule
& \multicolumn{2}{c}{Global $\ell_2$} & \multicolumn{2}{c}{\predictorN} \\
\cmidrule(lr){2-3}\cmidrule(lr){4-5}
& all & w/o & all & w/o \\
\midrule
1B     & $-0.56$ & $-0.36$ & $\mathbf{-0.92}$ & $\mathbf{-0.85}$ \\
3B     & $-0.17$ & $+0.13$ & $\mathbf{-0.72}$ & $\mathbf{-0.71}$ \\
Pooled & $-0.36$ & $-0.10$ & $\mathbf{-0.78}$ & $\mathbf{-0.74}$ \\
\bottomrule
\end{tabular}
\caption{Spearman $\rho$ between each predictor and $\Delta$ES under the
forget+retain relearning attack, over healthy checkpoints only (5 methods
$\times$ 3 splits $\times$ 2 models; $n\!=\!30$ pooled). Collapsed and noise
controls are excluded. More negative is better; ``w/o'' drops all \pruningN
checkpoints ($n\!=\!24$) to rule out circularity.}
\vspace{-12pt}
\label{tab:rank_corr}
\end{table}

%% file: latex/6_Conclusion.tex
\section{Conclusion}

We present a weight-selective view of relearning robustness: robust unlearning depends on which weights move, not distance alone. 
We introduce \predictorN, a training-free predictor of forget-critical and retain-sparing updates, and \pruningN, a direct pruning-based application of the same principle. 
Together, they show that forget-retain alignment provides a more reliable basis for predicting and improving relearning robustness.

%% file: latex/7_Limitation.tex
\section*{Limitations}

\looseness=-1 Our experiments cover TOFU, WMDP-cyber and MUSE-News across several model families and scales; broader benchmarks, multilingual data, and larger architectures would give a more complete picture. \predictorN also has limited resolution within dense unlearning methods: their updates receive scores an order of magnitude smaller than selective edits, so it separates dense from selective updates far more sharply than it ranks dense methods among themselves. \pruningN is instantiated as unstructured pruning; the same principle may extend to structured pruning, low-rank editing, and other parameter-efficient interventions. Studying relearning attacks also inevitably shows how easily unlearned knowledge can be recovered, which could inform adversaries seeking to restore hazardous content (e.g., WMDP); we use only public benchmarks and attack protocols, frame these attacks as tools for building more robust unlearning (\pruningN strengthens, not weakens, resistance), and release no model with restored hazardous capabilities.

%% file: latex/99_Appendix.tex
\clearpage
\appendix
\section*{Appendix}

The appendix expands three threads from the main text:
Appendix~\ref{app:fria} gives \predictorN's computational recipe and
the direction-blind perturbation control that motivates it.
Appendix~\ref{app:frip} reports \pruningN's implementation, evaluation
protocol, and ablations over scoring, mixing weight, and sparsity. Appendix~\ref{app:extended} provides the cross-family WMDP-cyber
extension and the per-split TOFU breakdowns supporting the averaged
main-text table.

\section{\predictorN: Forget--Retain Alignment Gap}
\label{app:fria}

This appendix complements \Cref{sec:frag} with \predictorN's full
computational recipe (\S\ref{app:fria_recipe}) and the
direction-blind perturbation control that motivates the
retain-penalty term (\S\ref{app:noise_ablation}).

\subsection{Computational Recipe}
\label{app:fria_recipe}

\paragraph{Module scope.}
\predictorN\ scores every linear projection inside each transformer
block (seven per block in Llama/Qwen): the four attention
projections \texttt{q\_proj}, \texttt{k\_proj}, \texttt{v\_proj},
\texttt{o\_proj}, and the three MLP projections \texttt{gate\_proj},
\texttt{up\_proj}, \texttt{down\_proj}. For other architectures the
predictor falls back to all \texttt{nn.Linear} modules; reported
results use this projection list.

\paragraph{Activation norms.}
Following the Wanda-style calibration of \citet{sun2024wanda},
forward hooks on each selected module accumulate per-input-channel
$\ell_2$ norms of the input activation:
\begin{equation}
x_j^{f,\ell} \;=\; \frac{1}{N_f}\!\sum_{i=1}^{N_f}
\bigl\lVert X^{(i),\ell}_{:,:,j} \bigr\rVert_{2},
\label{eq:actnorm}
\end{equation}
and analogously $x_j^{r,\ell}$. Both forward passes use $W_0$;
$W_u$ is read but never executed. Cosine similarity in
\Cref{eq:frag} is invariant to the per-channel norm convention.

\paragraph{Per-element importance and update tensors.}
For each layer $\ell$ with $W_0^\ell \in
\mathbb{R}^{d_{\text{out}} \times d_{\text{in}}}$, we broadcast
the channel norms along the output dimension:
\begin{equation}
\begin{aligned}
\mathcal{F}_{ij}^\ell &= |(W_0^\ell)_{ij}|\cdot
\frac{x_j^{f,\ell}}{x_j^{r,\ell}+\epsilon}, \\
\mathcal{R}_{ij}^\ell &= |(W_0^\ell)_{ij}|\cdot
\frac{x_j^{r,\ell}}{x_j^{f,\ell}+\epsilon},
\end{aligned}
\end{equation}
and form $D^\ell = (W_u^\ell - W_0^\ell)^{2}$.

\paragraph{Cross-layer aggregation.}
$A_f$ and $A_r$ are computed jointly across all selected layers
by streaming three inner products and three squared norms,
keeping memory at $O(d_{\text{out}}\!\times\!d_{\text{in}})$:
\begin{equation}
A_f \;=\;
\frac{\sum_\ell \langle \mathcal{F}^\ell, D^\ell \rangle}
{\sqrt{\sum_\ell \lVert \mathcal{F}^\ell \rVert^{2}}\;
 \sqrt{\sum_\ell \lVert D^\ell \rVert^{2}}}.
\end{equation}
This single global cosine, rather than per-layer cosines that are
then averaged, preserves the relative magnitude across layers so
updates concentrated in a few high-importance layers are rewarded
correctly.

\paragraph{Defaults.}
$\epsilon = 10^{-6}$, $\gamma = 1$, $N_f = N_r = 128$ calibration
sequences of length $256$, kept unchanged across all reported
results. The smaller calibration here vs.\ \pruningN's
$N\!=\!400$ (Table~\ref{tab:impl}) reflects that cosine alignment
is robust to per-channel norm noise, while \pruningN's per-row
top-$k$ thresholds require more stable estimates. Computation runs
in bf16 with fp32 accumulation; $W_u$ is read shard-by-shard from
safetensors so peak memory stays comparable to a single
transformer layer.

\paragraph{Compute cost.}
On a single A6000, end-to-end wall-clock is $25$~s on
LLaMA-3.2-1B (112 projections), $1$~min on 3B (196 projections),
and ${\sim}4$~min on Qwen-14B. The shortest relearning attack we
report (retain-1ep, batch $32$, lr$=$10$^{-5}$) costs ${\sim}30$~min
on 1B and over an hour on 14B per checkpoint, and yields only a
single post-attack ES point. \predictorN\ is ${\sim}60\times$
cheaper at 1B and ${\geq}15\times$ cheaper at 14B, and assigns a
continuous score from weights alone.

\subsection{Direction-blind Perturbation Ablation}
\label{app:noise_ablation}

The strongest test of \predictorN\ against $L_2$ is a perturbation
that performs no targeting at all: isotropic
$\mathcal{N}(0, \sigma^{2})$ noise added to every MLP weight of
Qwen2.5-14B-Instruct (seed 42). Attention layers are untouched,
no calibration data is used, and every weight is perturbed
equally. We choose $\sigma$ so that the resulting $L_2$
displacement straddles \pruningN's operating range:
$\sigma{=}0.002$ gives $L_2{\approx}202$, and $\sigma{=}0.004$
gives $L_2{\approx}404$, the latter matching \pruningN\ ($443.4$)
within $10\%$.

The noise rows of Table~\ref{tab:wmdp_tamper_combined} make the
distance/direction distinction sharp. At $\sigma{=}0.004$ the
noise edit not only matches \pruningN's $L_2$ but produces a more
negative $\Delta\mathrm{Acc}$ ($-0.027$ vs.\ $-0.012$); a predictor
that watches either signal would rank it as at least as robust as
\pruningN. The key observation is that pre-attack Acc is essentially
unchanged from Ref ($0.562$ vs.\ $0.583$) and MMLU actually
\emph{exceeds} \pruningN\ ($0.740$ vs.\ $0.668$) -- the noise did
not unlearn, so there is nothing for the attack to undo, and the
apparent robustness is vacuous. \predictorN\ correctly demotes both
noise settings to a flat $0.39$ ($\times 100$), well below
\pruningN's $9.20$ on the same panel. This is the failure mode
that motivates the retain term in Eq.~\eqref{eq:frag}: random
perturbations hit forget- and retain-critical weights with equal
intensity, so $A_f$ and $A_r$ are both large and roughly equal,
and the gap collapses. With $\gamma\!=\!0$, $A_f \approx 38\%$
exceeds \pruningN's $22\%$, the predictor would rank random
destruction as more robust.

\section{\pruningN: Forget--Retain Pruning}
\label{app:frip}

\subsection{Implementation Details}
\label{app:impl}
All experiments share the configuration in Table~\ref{tab:impl}.
Baseline-specific deviations from OpenUnlearning defaults are minimal:
RMU on WMDP targets \texttt{down\_proj} layers~5--7 with steering
coefficient 2; SP uses \texttt{mlp\_frac}$=$0.05,
\texttt{cos\_threshold}$=$0.5.

\begin{table}[h]
\centering
\footnotesize
\setlength{\tabcolsep}{5pt}
\renewcommand{\arraystretch}{1.15}
\begin{tabular}{@{}l l l@{}}
\toprule
\textbf{Component} & & \textbf{Setting} \\
\midrule
\multirow{5}{*}{\textbf{\pruningN}}
  & Sparsity         & 3\% (MLP only) \\
  & $\beta$          & 0.05 \\
  & Damage mode      & zero out ($W'{=}0$) \\
  & Calibration      & 400 seq $\times$ 256 tok \\
  & Precision        & bf16 ($\leq$3B), fp16 (7B+) \\
\midrule
\multirow{3}{*}{\textbf{Setup}}
  & Baselines        & OpenUnlearning defaults \\
  & Hardware         & 4$\times$RTX A6000 48\,GB \\
  & Seed             & 42 (all stages) \\
\bottomrule
\end{tabular}
\caption{\pruningN\ and infrastructure settings used across all
benchmarks. Attack and evaluation protocols are in
Appendix~\ref{app:eval}.}
\label{tab:impl}
\end{table}

\subsection{Evaluation Protocol}
\label{app:eval}

\paragraph{Forgetting and utility.}
On TOFU, forgetting is measured by Extraction
Strength~\citep{carlini2021extracting} (ES) as implemented in the
OpenUnlearning evaluator. For a sequence $y$ of length $|y|$
following prompt $x$, ES is one minus the normalized minimal prefix
length $k$ at which greedy continuation from $[x, y^{<k}]$ reproduces
the remaining suffix:
\begin{equation}
\mathrm{ES} = 1 - \frac{1}{|y|} \min_{k}\!\left\{ k \,\middle|\,
f\!\bigl([x, y^{<k}]; \theta\bigr) = y^{>k} \right\}.
\label{eq:es}
\end{equation}
$\mathrm{ES}=1$ indicates trivial extractability;
$\mathrm{ES}=0$ indicates the suffix cannot be recovered. We
report the per-split mean. Utility is the OpenUnlearning Model
Utility scalar (geometric mean of nine retain-side metrics). On
WMDP-cyber, ES is undefined for multiple-choice items, so we
substitute the benchmark's 4-way accuracy via
\texttt{lm-eval-harness}; the general-capability proxy is 5-shot
MMLU.

\paragraph{Three relearning attacks.}
We follow the ``Jogging the Memory'' threat
model~\citep{hu2024jogging}: after unlearning, the adversary
fine-tunes the released checkpoint for one epoch at lr$=$10$^{-5}$
with AdamW (8-bit on $\geq$7B). We instantiate the attacker with
three data sources to cover the realistic spectrum from
white-box leakage to mask-free recovery:
\begin{itemize}
\item \textbf{Retain attack}: adversary fine-tunes on the retain
split only. Tests whether normal continued use of the model surfaces
the forgotten content. This is the canonical Hu et al.\ protocol and
the default reported in the main text.
\item \textbf{Forget attack}: adversary has direct access to the
forget split itself (worst case: the leak that motivated unlearning
also reveals the forget set). Smallest split, $\sim$50 optimization
steps, but the strongest gradient signal toward the target content.
\item \textbf{Forget+retain attack}: full white-box adversary that
fine-tunes on the union, the most aggressive setting.
\end{itemize}
For each, we report post-attack ES, the change
$\Delta\mathrm{ES} = \mathrm{ES}_{\text{post}}-\mathrm{ES}_{\text{pre}}$,
and post-attack Utility. A robust unlearner should keep ES near its
unlearned value across all three attacks; a method that relies on
loss-surface suppression rather than knowledge removal will see
$\Delta\mathrm{ES}$ inflate sharply on the forget and
forget+retain attacks.

\paragraph{Averaging convention.}
Table~\ref{tab:tofu_attacks_avg_byMethod} averages over the three
forget splits ($1\%/5\%/10\%$) within each model. The split-wise
breakdown is reported in Appendix~\ref{app:tofu_full}.

\subsection{Ablation}
\label{app:ablation}

We ablate \pruningN's three design choices on LLaMA-3.2-1B / TOFU
forget10: the scoring rule (Table~\ref{tab:abl_scoring}), the
rank-space mixing weight $\beta$ (Table~\ref{tab:beta_sweep}), and
the sparsity $\rho$ (Table~\ref{tab:flp_sparsity_sweep_1B}).

\paragraph{Scoring rule.}
All three magnitude-driven baselines collapse utility to zero:
their top-scoring entries are the largest weights, which the
retain set also relies on, so the resulting low ES reflects
degraded output rather than targeted forgetting — their
\predictorN\ scores are correspondingly near zero. Only
\pruningN's rank-space combination of activation ratio $r$ and
$|W|$ ($\beta\!=\!0.05$) preserves utility above $0.4$ and earns
a positive \predictorN.
\input{Tables/ab1_scoring}

\paragraph{Mixing weight $\beta$.}
At fixed $3\%$ sparsity, $\beta$ traces a smooth, monotonic
utility/forgetting trade-off. We fix $\beta\!=\!0.05$ as the
smallest mixing weight that matches the Retain ES$_{\text{pre}}$
floor.
\input{Tables/beta_sweep}

\input{Tables/sparsity_sweep}
\paragraph{Sparsity $\rho$.}
At fixed $\beta\!=\!0.05$, sparsity sweeps a Pareto frontier
between forgetting and utility: ES$_{\text{pre}}$ falls
monotonically from $0.163$ at $0.5\%$ to $0.038$ at $10\%$ while
utility falls from $0.544$ to $0.143$. We adopt $3\%$ as the
headline operating point, the smallest sparsity that matches the
Retain ES$_{\text{pre}}$ floor ($0.062$ vs.\ $0.060$) while
preserving utility above $0.4$. The joint
$(\beta, \rho)\!=\!(0.05, 3\%)$ point is selected consistently by
the same criterion along both axes.
In practice, $\rho$ and $\beta$ should be increased only until the
target forgetting level is reached, since robustness gains beyond
that point are paid for in retain-side utility. Utility-critical
deployments should keep $\beta \le 0.05$ and use the smallest
sparsity that reaches the desired ES$_{\text{pre}}$.

\FloatBarrier
\section{Extended Results}
\label{app:extended}

\FloatBarrier
\subsection{Cross-Family on WMDP-cyber}
Table~\ref{tab:wmdp_tamper_combined} extends the WMDP-cyber
evaluation to Zephyr-7B-$\beta$~\cite{tunstall2024zephyr} and reports gradient-method
baselines; \predictorN\ and post-attack accuracy agree on the
ordering, while $\ell_2$ is inflated by direction-blind noise.
\input{Tables/wmdp_7B}

\subsection{Full TOFU Results}
\label{app:tofu_full}
Table~\ref{tab:tofu_full_per_split} expands
Table~\ref{tab:tofu_attacks_avg_byMethod} along two axes the main text
compresses: it separates the three relearning attacks instead of averaging
them, and it reports each forget split with its own retain-trained gold
reference. Two observations follow. First, under the forget+retain attack
\pruningN has the lowest attacked ES in every split at both scales; the
ordering is less stable under the weaker retain-only and forget-only attacks.
Second, the three attacks are not interchangeable: forget+retain
recovers the most for every method, and a checkpoint can appear robust under
the retain-only attack yet return most of the forgotten content once the
attacker also holds forget data---the same asymmetry the first-order argument
in \S\ref{sec:frag} predicts.

\input{Tables/full_attack_tofu}

\subsection{Relearning Attack Variants}
\label{app:attack_variants}
The main text fixes one attack configuration across its three attack sets.
Here we vary that configuration along four
axes on TOFU \texttt{forget10} with LLaMA-3.2-1B; every cell reports attacked
ES / utility. The RMU and NPO checkpoints in this section differ from those in
Table~\ref{tab:tofu_full_per_split}.
Not every setting is an effective attack: where no method
recovers meaningfully (all $\Delta$ES below $0.10$), nothing was taken back
from any of them and the setting says nothing about robustness. Excluding those
leaves eight effective variants, and \pruningN\ has the lowest attacked ES
among the unlearned methods in every one of them.

\paragraph{Learning rate.} Too small a step recovers nothing and too large a
one destroys the model, so only the middle of the range is informative
(Table~\ref{tab:attack_lr}).
\input{Tables/attack_lr}

\paragraph{Optimizer.} The attacker's optimizer matters more than its learning
rate: three of the five settings fail to constitute an attack at all
(Table~\ref{tab:attack_optim}).
\input{Tables/attack_optim}

\paragraph{Attack data.} Access to the forgotten examples is what makes an
attack strong; indirect substitutes recover far less
(Table~\ref{tab:attack_mixture}).
\input{Tables/attack_mixture}

\paragraph{Horizon.} Given enough epochs every method eventually gives the
knowledge back, so the question is how fast (Table~\ref{tab:attack_horizon}).
\input{Tables/attack_horizon}

\subsection{Matched Controls}
\label{app:matched}
Unlearning methods differ simultaneously in utility, forgetting depth,
sparsity, and update magnitude, so a raw comparison cannot attribute
\pruningN's robustness to selective placement rather than to one of these
confounds. We therefore address each in turn; the advantage survives all
four, which is what attributes it to \emph{where} the edit is placed.

\paragraph{Matched utility.} Sweeping each method's strength knob and comparing
only checkpoints of equal retain-side utility removes the possibility that
\pruningN\ simply trades utility for robustness (Table~\ref{tab:ctrl_utility}).
\input{Tables/ctrl_matched_utility}

\paragraph{Forgetting strength.} Comparing each method's pre-attack ES against
the gold level rules out the possibility that \pruningN\ merely forgets more
deeply to begin with: RMU forgets deeper still (0.056 vs.\ 0.062) and yet
recovers an order of magnitude more (Table~\ref{tab:ctrl_forget}).
\input{Tables/ctrl_matched_forget}

\paragraph{Matched sparsity.} Comparing \pruningN\ and SP at the same pruning
budget isolates the scoring rule from the amount of pruning
(Table~\ref{tab:ctrl_sparsity}).
\input{Tables/ctrl_matched_sparsity}

\paragraph{Matched update norm.} Comparing the two at near-identical
$\|\Delta W\|_2$ rules out displacement magnitude as the explanation
(Table~\ref{tab:ctrl_norm}).
\input{Tables/ctrl_matched_norm}
\FloatBarrier

\subsection{Predictor Design Choices}
\label{app:predictor_choices}
Two ingredients of \predictorN\ could have been chosen differently: the
importance term it is built on, and the weight-space quantity it competes
against. We check both.

\paragraph{Alternative importance measures.} \predictorN\ scores importance
from weight magnitude and input activation. Swapping that term for
gradient-, curvature-, or influence-based alternatives, and leaving the rest of
\predictorN\ untouched, weakens the predictor in every case
(Table~\ref{tab:importance_variants}).
\input{Tables/importance_variants}

\paragraph{Linear mode connectivity.} \citet{siddiqui2025dormant} pair
weight-space distance with a linear mode connectivity barrier in the vision
setting. Ported to TOFU, the barrier carries almost no signal
(Table~\ref{tab:lmc}).
\input{Tables/lmc}

\subsection{Cross-Benchmark Check}
\label{app:muse}
TOFU and WMDP-cyber differ in domain but share an extraction-style evaluation.
As a third setting we run MUSE-News~\cite{shi2025muse}, whose forget set is
natural news text rather than synthetic profiles or hazardous procedures, on
LLaMA-2-7B~\cite{touvron2023llama2} (Table~\ref{tab:muse}). The ordering among utility-preserving methods carries
over; broader benchmarks and architectures remain future work.
\input{Tables/muse_news}
\FloatBarrier

%% file: Tables/ab1_scoring.tex
\begin{table}[!htbp]
\centering
\footnotesize
\setlength{\tabcolsep}{2pt}
\renewcommand{\arraystretch}{1.05}
\begin{adjustbox}{max width=\columnwidth,center}
\begin{tabular}{@{}l|cc|ccc|cc@{}}
\toprule
\multirow{2}{*}{\textbf{Score}}
 & \multicolumn{2}{c|}{\textbf{Unlearned}}
 & \multicolumn{3}{c|}{\textbf{Retain Attack}}
 & \multicolumn{2}{c}{\textbf{Predictor}} \\
 & ES\,$\downarrow$ & U\,$\uparrow$
 & ES\,$\downarrow$ & $\Delta$ES\,$\downarrow$ & U\,$\uparrow$
 & L2\,$\uparrow$ & FRAG\,(\%)\,$\uparrow$ \\
\midrule
Ref                                       & 0.060 & 0.593 & 0.059 & -0.001 & 0.593 & ---   & ---  \\
\midrule
$|W|$ only$^{\dagger}$                    & 0.033 & 0.000 & 0.033 &  0.001 & 0.013 & 236.4 & -0.1 \\
$|W|\!\cdot\!\|X^f\|$$^{\dagger}$         & 0.033 & 0.000 & 0.033 &  0.000 & 0.008 & 214.9 & -0.1 \\
$|W|\!\cdot\!r$$^{\dagger}$               & 0.033 & 0.000 & 0.033 &  0.000 & 0.000 & 235.9 & +0.1 \\
\midrule
\textbf{FRP}\,($\beta{=}0.05$)                      & \colorbox{fst}{0.062} & \colorbox{fst}{0.405} & \colorbox{fst}{0.099} & \colorbox{fst}{0.037} & \colorbox{fst}{0.485} & \colorbox{fst}{111.3} & \colorbox{fst}{+2.1} \\
\bottomrule
\end{tabular}
\end{adjustbox}
\caption{Scoring ablation at fixed 3\% MLP sparsity on TOFU/LLaMA-3.2-1B
forget10. $^{\dagger}$Magnitude-driven scores collapse utility
(U\,$\leq$\,0.013 after attack) and are excluded from ranking.}
\label{tab:abl_scoring}
\end{table}

%% file: Tables/beta_sweep.tex
\begin{table}[!htbp]
\centering
\footnotesize
\setlength{\tabcolsep}{2pt}
\renewcommand{\arraystretch}{1.05}
\begin{adjustbox}{max width=\columnwidth,center}
\begin{tabular}{@{}c|cc|ccc|cc@{}}
\toprule
\multirow{2}{*}{$\beta$}
 & \multicolumn{2}{c|}{\textbf{Unlearned}}
 & \multicolumn{3}{c|}{\textbf{Retain Attack}}
 & \multicolumn{2}{c}{\textbf{Predictor}} \\
 & ES\,$\downarrow$ & U\,$\uparrow$
 & ES\,$\downarrow$ & $\Delta$ES\,$\downarrow$ & U\,$\uparrow$
 & L2\,$\uparrow$ & \predictorN\,(\%)\,$\uparrow$ \\
\midrule
Ref                  & 0.060 & 0.593 & 0.059 & $-$0.001 & 0.593 & ---   & ---  \\
\midrule
0.00                 & 0.086 & 0.481 & 0.121 & +0.035 & 0.527 &  89.4 & +1.9 \\
0.01                 & 0.082 & 0.464 & 0.112 & +0.031 & 0.513 &  94.1 & +1.9 \\
\textbf{0.05}$^{\star}$ & \colorbox{fst}{0.062} & 0.408 & 0.100 & +0.037 & 0.482 & 111.3 & +2.1 \\
0.10                 & 0.053 & 0.359 & 0.080 & +0.027 & 0.443 & 127.0 & +2.1 \\
0.25                 & 0.046 & 0.295 & 0.078 & +0.032 & 0.393 & 147.8 & +2.2 \\
0.50                 & 0.045 & 0.220 & 0.070 & +0.025 & 0.324 & 162.7 & +2.2 \\
1.00                 & 0.041 & 0.167 & 0.065 & +0.024 & 0.301 & 177.1 & +2.2 \\
\bottomrule
\end{tabular}
\end{adjustbox}
\caption{$\beta$ sweep at 3\% MLP sparsity on TOFU/LLaMA-3.2-1B
forget10. $^{\star}$Headline, chosen as the smallest $\beta$ matching
the Retain ES$_{\text{pre}}$ floor.}
\label{tab:beta_sweep}
\end{table}

%% file: Tables/sparsity_sweep.tex
\begin{table}[!htbp]
\centering
\footnotesize
\setlength{\tabcolsep}{2pt}
\renewcommand{\arraystretch}{1.05}
\begin{adjustbox}{max width=\columnwidth,center}
\begin{tabular}{@{}c|cc|ccc|cc@{}}
\toprule
\multirow{2}{*}{\textbf{sp\,(\%)}}
 & \multicolumn{2}{c|}{\textbf{Unlearned}}
 & \multicolumn{3}{c|}{\textbf{Retain Attack}}
 & \multicolumn{2}{c}{\textbf{Predictor}} \\
 & ES\,$\downarrow$ & U\,$\uparrow$
 & ES\,$\downarrow$ & $\Delta$ES\,$\downarrow$ & U\,$\uparrow$
 & L2\,$\uparrow$ & FRAG\,(\%)\,$\uparrow$ \\
\midrule
Ref            & 0.060 & 0.593 & 0.059 & -0.001 & 0.593 & ---    & ---  \\
\midrule
0.5            & 0.163 & 0.544 & 0.219 & 0.055  & 0.562 & 60.95  & +1.5 \\
1.0            & 0.107 & 0.520 & 0.147 & 0.041  & 0.548 & 77.83  & +1.7 \\
1.5            & 0.090 & 0.477 & 0.126 & 0.036  & 0.521 & 89.07  & +1.8 \\
2.0            & 0.078 & 0.461 & 0.114 & 0.036  & 0.508 & 97.65  & +1.9 \\
2.5            & 0.070 & 0.431 & 0.108 & 0.038  & 0.497 & 105.15 & +1.9 \\
{3.0}   & {0.062} & {0.408} & {0.100} & {0.037} & {0.482} & {111.34} & {+2.1} \\
4.0            & 0.055 & 0.375 & 0.084 & 0.029  & 0.451 & 122.77 & +2.1 \\
5.0            & 0.051 & 0.316 & 0.076 & 0.025  & 0.421 & 133.59 & +2.1 \\
7.0            & 0.045 & 0.245 & 0.068 & 0.023  & 0.359 & 152.75 & +2.2 \\
10.0           & 0.038 & 0.143 & 0.064 & 0.026  & 0.263 & 177.58 & +2.4 \\
\bottomrule
\end{tabular}
\end{adjustbox}
\caption{LLaMA-3.2-1B FRP sparsity sweep on TOFU forget10.
{3.0\%} is the headline operating point; \textit{Ref} is
the retain90 gold model.}
\label{tab:flp_sparsity_sweep_1B}
\end{table}

%% file: Tables/wmdp_7B.tex
\begin{table}[!htbp]
\centering
\footnotesize
\setlength{\tabcolsep}{2pt}
\renewcommand{\arraystretch}{1.05}
\begin{adjustbox}{max width=\columnwidth,center}
\begin{tabular}{@{}l|cc|cc|cc@{}}
\toprule
\multirow{2}{*}{\textbf{Method}}
 & \multicolumn{2}{c|}{\textbf{Unlearned}}
 & \multicolumn{2}{c|}{\textbf{Retain Attack}}
 & \multicolumn{2}{c}{\textbf{Predictor}} \\
 & Acc\,$\downarrow$ & MMLU\,$\uparrow$
 & Acc\,$\downarrow$ & MMLU\,$\uparrow$
 & L2\,$\uparrow$ & \predictorN\,(\%)\,$\uparrow$ \\
\midrule\midrule
\multicolumn{7}{c}{\textit{Zephyr-7B-$\beta$}} \\
\midrule
Ref                  & 0.446 & 0.586 & 0.430 & 0.577 & ---   & ---   \\
\midrule
GA$^{\dagger}$       & 0.243 & 0.247 & 0.254 & 0.256 &  25.7 & 0.063 \\
GradDiff$^{\dagger}$ & 0.246 & 0.255 & 0.246 & 0.255 &  27.5 & 0.052 \\
NPO$^{\dagger}$      & 0.245 & 0.251 & 0.246 & 0.255 &  26.8 & 0.182 \\
RMU                  & \colorbox{fst}{0.270} & \colorbox{fst}{0.574} & \colorbox{thd}{0.424} & \colorbox{fst}{0.577} &   4.7 & \colorbox{thd}{0.105} \\
SP                   & \colorbox{thd}{0.414} & \colorbox{sed}{0.572} & \colorbox{sed}{0.407} & \colorbox{sed}{0.561} & \colorbox{fst}{52.1}  & \colorbox{sed}{2.148} \\
\midrule
\textbf{\pruningN}
                     & \colorbox{sed}{0.344} & \colorbox{thd}{0.533} & \colorbox{fst}{0.376} & \colorbox{thd}{0.530} & \colorbox{sed}{49.9}  & \colorbox{fst}{6.311} \\
\midrule\midrule
\multicolumn{7}{c}{\textit{Qwen2.5-14B-Instruct}} \\
\midrule
Ref                  & 0.583 & 0.788 & 0.581 & 0.792 & ---   & ---   \\
\midrule
GA$^{\dagger}$       & 0.255 & 0.270 & 0.264 & 0.241 & 288.5 & 0.054 \\
GradDiff$^{\dagger}$ & 0.245 & 0.726 & 0.275 & 0.693 & 507.1 & 0.004 \\
RMU                  & \colorbox{thd}{0.479} & \colorbox{fst}{0.782} & 0.552                 & \colorbox{fst}{0.791} &  26.3 & 0.185 \\
SP                   & \colorbox{sed}{0.515} & \colorbox{thd}{0.764} & \colorbox{sed}{0.513} & \colorbox{thd}{0.771} & \colorbox{fst}{464.1} & \colorbox{sed}{3.149} \\
\midrule
Noise $\sigma{=}.002$ & 0.582 & \colorbox{sed}{0.778} & 0.557 & \colorbox{sed}{0.782} & 201.9 & \colorbox{thd}{0.394} \\
Noise $\sigma{=}.004$ & 0.562 & 0.740 & \colorbox{thd}{0.535} & 0.752 & \colorbox{thd}{403.8} & \colorbox{thd}{0.394} \\
\midrule
\textbf{\pruningN}
                     & \colorbox{fst}{0.429} & 0.668 & \colorbox{fst}{0.417} & 0.707 & \colorbox{sed}{443.4} & \colorbox{fst}{9.198} \\
\bottomrule
\end{tabular}
\end{adjustbox}
\caption{Cross-family validation on WMDP-cyber
(Zephyr-7B-$\beta$, Qwen2.5-14B-Instruct); MMLU as the
general-capability proxy. ${}^{\dagger}$ marks degenerate
baselines excluded from ranking.}
\label{tab:wmdp_tamper_combined}
\end{table}

%% file: Tables/full_attack_tofu.tex
\begin{table*}[!htbp]
\centering
\footnotesize
\setlength{\tabcolsep}{3pt}
\renewcommand{\arraystretch}{1.0}
\begin{adjustbox}{max width=\textwidth,center}
\begin{tabular}{@{}l|cc|ccc|ccc|ccc|cc@{}}
\toprule
\multirow{2}{*}{\textbf{Method}}
 & \multicolumn{2}{c|}{\textbf{Unlearned}}
 & \multicolumn{3}{c|}{\textbf{Retain Attack}}
 & \multicolumn{3}{c|}{\textbf{Forget Attack}}
 & \multicolumn{3}{c|}{\textbf{Forget+Retain Attack}}
 & \multicolumn{2}{c}{\textbf{Predictor}} \\
 & ES\,$\downarrow$ & U\,$\uparrow$
 & ES\,$\downarrow$ & $\Delta$ES\,$\downarrow$ & U\,$\uparrow$
 & ES\,$\downarrow$ & $\Delta$ES\,$\downarrow$ & U\,$\uparrow$
 & ES\,$\downarrow$ & $\Delta$ES\,$\downarrow$ & U\,$\uparrow$
 & L2\,$\uparrow$ & \predictorN(\%)\,$\uparrow$ \\
\midrule
\multicolumn{14}{c}{\textbf{TOFU forget01}} \\
\midrule
\multicolumn{14}{c}{\textit{LLaMA-3.2-1B}} \\
\midrule
Retain   & 0.069 & 0.598 & 0.067 & -0.002 & 0.597 & 0.096 & +0.027 & 0.593 & 0.073 & +0.003 & 0.601 & 2.781 & - \\
\midrule
GA                        & 0.187 & \colorbox{sed}{0.594} & 0.185 & \colorbox{sed}{-0.002} & \colorbox{thd}{0.601} & 0.309 & +0.122 & \colorbox{thd}{0.591} & 0.298 & +0.112 & \colorbox{sed}{0.600} & \colorbox{thd}{0.361} & -0.002 \\
GradDiff                  & 0.178 & \colorbox{thd}{0.587} & 0.240 & +0.062 & \colorbox{sed}{0.602} & 0.304 & +0.126 & \colorbox{fst}{0.601} & 0.348 & +0.170 & 0.599 & 0.314 & \colorbox{thd}{+0.003} \\
NPO                       & 0.181 & \colorbox{fst}{0.595} & \colorbox{thd}{0.148} & \colorbox{fst}{-0.033} & 0.600 & \colorbox{thd}{0.292} & \colorbox{thd}{+0.110} & \colorbox{sed}{0.593} & \colorbox{thd}{0.260} & \colorbox{thd}{+0.079} & \colorbox{thd}{0.600} & 0.353 & -0.002 \\
RMU                       & \colorbox{thd}{0.150} & 0.556 & 0.433 & +0.283 & \colorbox{fst}{0.604} & 0.485 & +0.335 & 0.587 & 0.792 & +0.642 & \colorbox{fst}{0.601} & 0.284 & +0.003 \\
SP                        & \colorbox{sed}{0.079} & 0.456 & \colorbox{sed}{0.104} & +0.025 & 0.495 & \colorbox{sed}{0.151} & \colorbox{sed}{+0.073} & 0.465 & \colorbox{sed}{0.150} & \colorbox{sed}{+0.072} & 0.494 & \colorbox{fst}{121.7} & \colorbox{sed}{+0.689} \\
\textbf{\pruningN}($\beta{=}0.05$)  & \colorbox{fst}{0.036} & 0.344 & \colorbox{fst}{0.052} & \colorbox{thd}{+0.015} & 0.445 & \colorbox{fst}{0.041} & \colorbox{fst}{+0.005} & 0.369 & \colorbox{fst}{0.071} & \colorbox{fst}{+0.035} & 0.439 & \colorbox{sed}{111.8} & \colorbox{fst}{+5.238} \\
\midrule
\multicolumn{14}{c}{\textit{LLaMA-3.2-3B}} \\
\midrule
Retain   & 0.067 & 0.663 & 0.088 & +0.021 & 0.663 & 0.099 & +0.033 & 0.658 & 0.094 & +0.028 & 0.663 & 4.760 & - \\
\midrule
GA                        & 0.237 & \colorbox{fst}{0.667} & 0.244 & \colorbox{fst}{+0.007} & \colorbox{sed}{0.659} & 0.402 & \colorbox{thd}{+0.165} & \colorbox{sed}{0.664} & 0.417 & +0.180 & \colorbox{sed}{0.658} & 0.576 & +0.002 \\
GradDiff                  & 0.318 & \colorbox{sed}{0.661} & 0.357 & +0.039 & \colorbox{thd}{0.657} & 0.501 & +0.184 & \colorbox{fst}{0.668} & 0.484 & \colorbox{thd}{+0.166} & \colorbox{thd}{0.658} & 0.503 & +0.008 \\
NPO                       & \colorbox{thd}{0.119} & 0.655 & \colorbox{sed}{0.138} & \colorbox{thd}{+0.019} & 0.657 & \colorbox{sed}{0.182} & \colorbox{sed}{+0.063} & 0.658 & \colorbox{sed}{0.187} & \colorbox{sed}{+0.068} & 0.653 & \colorbox{thd}{1.032} & +0.004 \\
RMU                       & \colorbox{sed}{0.075} & \colorbox{thd}{0.660} & 0.237 & +0.162 & \colorbox{fst}{0.663} & \colorbox{thd}{0.240} & +0.165 & \colorbox{thd}{0.662} & 0.721 & +0.646 & \colorbox{fst}{0.661} & 0.877 & \colorbox{thd}{+0.016} \\
SP                        & 0.123 & 0.588 & \colorbox{thd}{0.214} & +0.091 & 0.618 & 0.351 & +0.228 & 0.592 & \colorbox{thd}{0.330} & +0.206 & 0.620 & \colorbox{fst}{192.3} & \colorbox{sed}{+1.628} \\
\textbf{\pruningN}($\beta{=}0.05$)  & \colorbox{fst}{0.046} & 0.458 & \colorbox{fst}{0.056} & \colorbox{sed}{+0.010} & 0.540 & \colorbox{fst}{0.065} & \colorbox{fst}{+0.019} & 0.473 & \colorbox{fst}{0.083} & \colorbox{fst}{+0.037} & 0.545 & \colorbox{sed}{182.3} & \colorbox{fst}{+5.782} \\
\midrule\midrule
\multicolumn{14}{c}{\textbf{TOFU forget05}} \\
\midrule
\multicolumn{14}{c}{\textit{LLaMA-3.2-1B}} \\
\midrule
Retain   & 0.063 & 0.598 & 0.063 & -0.000 & 0.600 & 0.076 & +0.013 & 0.580 & 0.070 & +0.007 & 0.602 & 2.975 & - \\
\midrule
GA                        & \colorbox{fst}{0.039} & 0.002 & 0.276 & +0.237 & 0.596 & \colorbox{fst}{0.045} & \colorbox{fst}{+0.006} & 0.078 & 0.418 & +0.379 & 0.592 & \colorbox{thd}{0.936} & +0.003 \\
GradDiff                  & 0.108 & \colorbox{thd}{0.467} & 0.258 & +0.150 & \colorbox{fst}{0.602} & 0.174 & +0.066 & \colorbox{fst}{0.578} & 0.462 & +0.354 & \colorbox{fst}{0.601} & 0.614 & +0.003 \\
NPO                       & \colorbox{thd}{0.101} & 0.464 & \colorbox{thd}{0.212} & \colorbox{thd}{+0.111} & \colorbox{thd}{0.600} & \colorbox{thd}{0.135} & \colorbox{thd}{+0.033} & 0.504 & \colorbox{thd}{0.339} & \colorbox{thd}{+0.237} & \colorbox{sed}{0.600} & 0.819 & +0.003 \\
RMU                       & 0.108 & \colorbox{fst}{0.551} & 0.512 & +0.404 & \colorbox{sed}{0.602} & 0.356 & +0.248 & \colorbox{sed}{0.577} & 0.733 & +0.625 & \colorbox{thd}{0.600} & 0.749 & \colorbox{thd}{+0.006} \\
SP                        & 0.163 & \colorbox{sed}{0.504} & \colorbox{sed}{0.210} & \colorbox{sed}{+0.047} & 0.535 & 0.234 & +0.072 & \colorbox{thd}{0.513} & \colorbox{sed}{0.285} & \colorbox{sed}{+0.122} & 0.532 & \colorbox{fst}{119.8} & \colorbox{sed}{+0.222} \\
\textbf{\pruningN}($\beta{=}0.05$)  & \colorbox{sed}{0.067} & 0.399 & \colorbox{fst}{0.086} & \colorbox{fst}{+0.018} & 0.464 & \colorbox{sed}{0.081} & \colorbox{sed}{+0.013} & 0.420 & \colorbox{fst}{0.120} & \colorbox{fst}{+0.053} & 0.460 & \colorbox{sed}{111.5} & \colorbox{fst}{+2.156} \\
\midrule
\multicolumn{14}{c}{\textit{LLaMA-3.2-3B}} \\
\midrule
Retain   & 0.061 & 0.660 & 0.061 & -0.001 & 0.658 & 0.082 & +0.021 & 0.649 & 0.077 & +0.016 & 0.651 & 4.994 & - \\
\midrule
GA                        & 0.091 & 0.484 & 0.316 & +0.224 & \colorbox{sed}{0.664} & 0.165 & +0.074 & 0.584 & \colorbox{thd}{0.491} & +0.400 & \colorbox{fst}{0.669} & 1.473 & +0.002 \\
GradDiff                  & 0.162 & 0.562 & 0.385 & +0.223 & 0.659 & 0.384 & +0.222 & \colorbox{sed}{0.643} & 0.543 & +0.382 & 0.659 & 1.046 & \colorbox{thd}{+0.003} \\
NPO                       & \colorbox{sed}{0.070} & \colorbox{sed}{0.662} & \colorbox{sed}{0.101} & \colorbox{sed}{+0.031} & \colorbox{thd}{0.662} & \colorbox{thd}{0.138} & \colorbox{thd}{+0.068} & \colorbox{thd}{0.640} & \colorbox{sed}{0.259} & \colorbox{sed}{+0.189} & \colorbox{sed}{0.667} & 2.024 & +0.001 \\
RMU                       & \colorbox{fst}{0.054} & \colorbox{fst}{0.665} & \colorbox{fst}{0.073} & \colorbox{fst}{+0.019} & \colorbox{fst}{0.665} & \colorbox{fst}{0.102} & \colorbox{fst}{+0.048} & \colorbox{fst}{0.664} & 0.733 & +0.678 & \colorbox{thd}{0.663} & \colorbox{thd}{2.272} & +0.001 \\
SP                        & 0.219 & \colorbox{thd}{0.590} & 0.341 & +0.122 & 0.611 & 0.486 & +0.267 & 0.600 & 0.494 & \colorbox{thd}{+0.275} & 0.618 & \colorbox{fst}{191.7} & \colorbox{sed}{+0.615} \\
\textbf{\pruningN}($\beta{=}0.05$)  & \colorbox{thd}{0.078} & 0.512 & \colorbox{thd}{0.116} & \colorbox{thd}{+0.038} & 0.579 & \colorbox{sed}{0.132} & \colorbox{sed}{+0.054} & 0.539 & \colorbox{fst}{0.183} & \colorbox{fst}{+0.105} & 0.575 & \colorbox{sed}{182.3} & \colorbox{fst}{+2.537} \\
\midrule\midrule
\multicolumn{14}{c}{\textbf{TOFU forget10}} \\
\midrule
\multicolumn{14}{c}{\textit{LLaMA-3.2-1B}} \\
\midrule
Retain   & 0.060 & 0.593 & 0.059 & -0.001 & 0.593 & 0.071 & +0.012 & 0.568 & 0.070 & +0.010 & 0.589 & 3.255 & - \\
\midrule
GA                        & \colorbox{fst}{0.033} & 0.000 & 0.392 & +0.359 & 0.596 & \colorbox{sed}{0.104} & \colorbox{thd}{+0.072} & 0.407 & 0.486 & +0.453 & \colorbox{fst}{0.601} & \colorbox{thd}{1.311} & \colorbox{thd}{+0.005} \\
GradDiff                  & 0.082 & \colorbox{thd}{0.442} & \colorbox{thd}{0.271} & \colorbox{thd}{+0.189} & \colorbox{sed}{0.602} & 0.315 & +0.233 & \colorbox{fst}{0.595} & 0.512 & +0.430 & 0.600 & 0.773 & +0.003 \\
NPO                       & 0.096 & 0.402 & 0.285 & +0.190 & \colorbox{fst}{0.609} & \colorbox{thd}{0.164} & \colorbox{sed}{+0.069} & \colorbox{thd}{0.544} & \colorbox{thd}{0.396} & \colorbox{thd}{+0.300} & \colorbox{sed}{0.601} & 1.065 & +0.001 \\
RMU                       & \colorbox{sed}{0.056} & \colorbox{fst}{0.577} & 0.291 & +0.235 & \colorbox{thd}{0.599} & 0.420 & +0.363 & \colorbox{sed}{0.574} & 0.701 & +0.644 & \colorbox{thd}{0.601} & 1.264 & +0.002 \\
SP                        & 0.131 & \colorbox{sed}{0.499} & \colorbox{sed}{0.200} & \colorbox{sed}{+0.069} & 0.533 & 0.213 & +0.082 & 0.513 & \colorbox{sed}{0.275} & \colorbox{sed}{+0.144} & 0.535 & \colorbox{fst}{120.1} & \colorbox{sed}{+0.269} \\
\textbf{\pruningN}($\beta{=}0.05$)  & \colorbox{thd}{0.062} & 0.408 & \colorbox{fst}{0.100} & \colorbox{fst}{+0.037} & 0.482 & \colorbox{fst}{0.090} & \colorbox{fst}{+0.027} & 0.432 & \colorbox{fst}{0.121} & \colorbox{fst}{+0.059} & 0.488 & \colorbox{sed}{111.3} & \colorbox{fst}{+2.083} \\
\midrule
\multicolumn{14}{c}{\textit{LLaMA-3.2-3B}} \\
\midrule
Retain   & 0.063 & 0.651 & 0.064 & +0.001 & 0.645 & 0.075 & +0.012 & 0.611 & 0.075 & +0.012 & 0.638 & 5.282 & - \\
\midrule
GA                        & \colorbox{fst}{0.033} & 0.000 & 0.433 & +0.401 & \colorbox{fst}{0.689} & \colorbox{sed}{0.098} & \colorbox{thd}{+0.066} & 0.395 & 0.581 & +0.549 & \colorbox{fst}{0.692} & 2.041 & +0.003 \\
GradDiff                  & 0.107 & 0.529 & 0.397 & +0.290 & 0.656 & 0.478 & +0.371 & \colorbox{sed}{0.652} & 0.632 & +0.525 & 0.663 & 1.257 & \colorbox{thd}{+0.003} \\
NPO                       & \colorbox{thd}{0.057} & \colorbox{fst}{0.674} & \colorbox{sed}{0.085} & \colorbox{sed}{+0.027} & \colorbox{sed}{0.677} & 0.204 & +0.147 & \colorbox{thd}{0.650} & \colorbox{sed}{0.396} & \colorbox{thd}{+0.339} & \colorbox{sed}{0.672} & 2.559 & +0.000 \\
RMU                       & \colorbox{sed}{0.034} & \colorbox{sed}{0.667} & \colorbox{fst}{0.058} & \colorbox{fst}{+0.024} & \colorbox{thd}{0.665} & \colorbox{fst}{0.075} & \colorbox{fst}{+0.041} & \colorbox{fst}{0.664} & 0.810 & +0.776 & \colorbox{thd}{0.669} & \colorbox{thd}{3.082} & +0.003 \\
SP                        & 0.217 & \colorbox{thd}{0.587} & 0.341 & +0.123 & 0.612 & 0.456 & +0.238 & 0.600 & \colorbox{thd}{0.507} & \colorbox{sed}{+0.289} & 0.617 & \colorbox{fst}{191.8} & \colorbox{sed}{+0.580} \\
\textbf{\pruningN}($\beta{=}0.05$)  & 0.081 & 0.528 & \colorbox{thd}{0.117} & \colorbox{thd}{+0.036} & 0.588 & \colorbox{thd}{0.125} & \colorbox{sed}{+0.044} & 0.553 & \colorbox{fst}{0.175} & \colorbox{fst}{+0.094} & 0.589 & \colorbox{sed}{181.9} & \colorbox{fst}{+2.295} \\
\bottomrule
\end{tabular}
\end{adjustbox}
\caption{Per-method tamper resistance on TOFU forget01/05/10. \textit{Retain} is the gold reference (unranked).}
\label{tab:tofu_full_per_split}
\end{table*}

%% file: Tables/attack_lr.tex
\begin{table}[H]
\centering
\small
\setlength{\tabcolsep}{3pt}
\begin{tabular}{lcccc}
\toprule
lr & gold & RMU & NPO & \pruningN \\
\midrule
pre-attack        & .060/.593 & .035/.581 & .085/.568 & .062/.405 \\
$5\mathrm{e}{-6}$ & .062/.590 & .047/.584 & .104/.565 & .095/.467 \\
$1\mathrm{e}{-5}$ & .068/.588 & .461/.587 & .137/.576 & \textbf{.120}/.484 \\
$2\mathrm{e}{-5}$ & .103/.582 & .611/.592 & .259/.590 & \textbf{.178}/.499 \\
$5\mathrm{e}{-5}$ & .135/.541 & .358/.550 & .270/.550 & \textbf{.207}/.474 \\
$1\mathrm{e}{-4}$ & .109/.446 & .134/.445 & .134/.442 & \textbf{.129}/.337 \\
\bottomrule
\end{tabular}
\caption{Relearning attack across learning rates (forget+retain attack set,
1 epoch, AdamW), as attacked ES / utility; lower ES means less recovery.
Recovery is negligible at $5\mathrm{e}{-6}$; from $1\mathrm{e}{-5}$ to
$5\mathrm{e}{-5}$, \pruningN\ has lower attacked ES than NPO and RMU. At
$1\mathrm{e}{-4}$ every method loses substantial utility.}
\label{tab:attack_lr}
\end{table}

%% file: Tables/attack_optim.tex
\begin{table}[H]
\centering
\small
\setlength{\tabcolsep}{3pt}
\begin{tabular}{lcccc}
\toprule
Optimizer & gold & RMU & NPO & \pruningN \\
\midrule
pre-attack     & .060/.593 & .035/.581 & .085/.568 & .062/.405 \\
AdamW          & .109/.446 & .134/.445 & .134/.442 & .129/.337 \\
SGD            & .059/.592 & .035/.581 & .091/.571 & .064/.416 \\
SGD+mom.\      & .060/.592 & .035/.574 & .096/.571 & .076/.439 \\
Adafactor      & .083/.297 & .088/.274 & .086/.280 & .079/.108 \\
Adagrad        & .083/.583 & .597/.590 & .208/.586 & \textbf{.137}/.489 \\
\bottomrule
\end{tabular}
\caption{Relearning attack across optimizers (forget+retain attack set,
1 epoch, lr $1\mathrm{e}{-4}$), as attacked ES / utility. SGD recovers
nothing---every method, including the retain-trained gold model, stays at its
pre-attack ES---and Adafactor collapses utility for all methods, so neither
probes robustness. ``SGD+mom.'' uses momentum 0.9. Adagrad is the only additional effective attack, and there
\pruningN\ remains far less recovered than RMU and NPO.}
\label{tab:attack_optim}
\end{table}

%% file: Tables/attack_mixture.tex
\begin{table}[H]
\centering
\small
\setlength{\tabcolsep}{3pt}
\begin{tabular}{lcccc}
\toprule
Attack set & gold & RMU & NPO & \pruningN \\
\midrule
pre-attack       & .060/.593 & .035/.581 & .085/.568 & .062/.405 \\
retain only      & .059/.594 & .039/.598 & .097/.585 & .099/.485 \\
forget only      & .066/.573 & .038/.576 & .093/.535 & .090/.437 \\
forget+retain    & .068/.588 & .461/.587 & .137/.576 & \textbf{.120}/.484 \\
paraphrase       & .065/.572 & .037/.576 & .084/.520 & .075/.425 \\
50/50 mix        & .067/.568 & .087/.567 & .111/.544 & .101/.452 \\
forget+general   & .069/.642 & .047/.597 & .099/.575 & .097/.454 \\
\bottomrule
\end{tabular}
\caption{Relearning attack across attack-data mixtures (1 epoch, lr
$1\mathrm{e}{-5}$, AdamW), as attacked ES / utility. Single-source and indirect
mixtures recover little; ``paraphrase'' rewrites the forget set and ``50/50 mix'' balances forget and retain; forget+retain is the strongest attack, and it is there
that \pruningN\ shows the lowest attacked ES among the unlearned models.}
\label{tab:attack_mixture}
\end{table}

%% file: Tables/attack_horizon.tex
\begin{table}[H]
\centering
\small
\setlength{\tabcolsep}{3pt}
\begin{tabular}{lcccc}
\toprule
Horizon & gold & RMU & NPO & \pruningN \\
\midrule
pre-attack & .060/.593 & .035/.581 & .085/.568 & .062/.405 \\
1 ep       & .068/.588 & .461/.587 & .137/.576 & \textbf{.120}/.484 \\
2 ep       & .084/.588 & .649/.594 & .200/.584 & \textbf{.161}/.496 \\
3 ep       & .097/.586 & .792/.596 & .255/.584 & \textbf{.203}/.502 \\
5 ep       & .126/.585 & .936/.596 & .404/.585 & \textbf{.318}/.508 \\
10 ep      & .365/.575 & .999/.587 & .820/.566 & \textbf{.752}/.504 \\
\bottomrule
\end{tabular}
\caption{Relearning attack across fine-tuning horizons (forget+retain attack
set, lr $1\mathrm{e}{-5}$, AdamW), as attacked ES / utility. \pruningN\ stays
less recovered than NPO and RMU at every horizon; after 10 epochs RMU reaches
.999 and NPO .820 while \pruningN\ remains at .752.}
\label{tab:attack_horizon}
\end{table}

%% file: Tables/ctrl_matched_utility.tex
\begin{table}[H]
\centering
\small
\setlength{\tabcolsep}{6pt}
\begin{tabular}{lccc}
\toprule
& \texttt{forget01} & \texttt{forget05} & \texttt{forget10} \\
\midrule
\multicolumn{4}{l}{\emph{utility} $\approx 0.52$} \\
\quad RMU       & .486 & .279 & .304 \\
\quad \pruningN & \textbf{.101} & \textbf{.086} & \textbf{.156} \\
\addlinespace[2pt]
\multicolumn{4}{l}{\emph{utility} $\approx 0.45$} \\
\quad RMU       & --- & .263 & .228 \\
\quad \pruningN & \textbf{.052} & \textbf{.056} & \textbf{.075} \\
\addlinespace[2pt]
\multicolumn{4}{l}{\emph{utility} $\approx 0.38$} \\
\quad RMU       & --- & .257 & --- \\
\quad \pruningN & \textbf{.035} & \textbf{.053} & \textbf{.058} \\
\bottomrule
\end{tabular}
\caption{$\Delta$ES at matched utility (TOFU-1B, forget+retain attack,
1 epoch). Each method's strength knob is swept---steering coefficient for RMU,
sparsity for \pruningN---and checkpoints are grouped into utility bands;
``---'' marks a band with no RMU checkpoint. Within every band where both
methods appear, utilities differ by at most $0.01$ or \pruningN's is higher,
and \pruningN\ recovers $1.9$--$4.9\times$ less.}
\label{tab:ctrl_utility}
\end{table}

%% file: Tables/ctrl_matched_forget.tex
\begin{table}[H]
\centering
\small
\setlength{\tabcolsep}{6pt}
\begin{tabular}{lccc}
\toprule
Method & pre-ES & pre-utility & $\Delta$ES \\
\midrule
Retain (gold) & 0.060 & 0.593 & +0.008 \\
\midrule
RMU        & 0.056 & 0.577 & +0.644 \\
GradDiff   & 0.082 & 0.442 & +0.430 \\
NPO        & 0.096 & 0.402 & +0.300 \\
SP         & 0.131 & 0.499 & +0.144 \\
\pruningN  & 0.062 & 0.408 & \textbf{+0.059} \\
\bottomrule
\end{tabular}
\caption{Forgetting strength on \texttt{forget10}, compared against the gold
pre-attack ES (0.060). \pruningN\ lands closest to the gold level (0.062) and
still recovers the least; RMU forgets even more deeply (0.056) yet recovers
about ten times as much.}
\label{tab:ctrl_forget}
\end{table}

%% file: Tables/ctrl_matched_sparsity.tex
\begin{table}[H]
\centering
\small
\setlength{\tabcolsep}{6pt}
\begin{tabular}{llccc}
\toprule
Budget & Method & pre-ES & pre-utility & $\Delta$ES \\
\midrule
2.5\% & SP        & 0.274 & 0.539 & +0.173 \\
2.5\% & \pruningN & \textbf{0.068} & 0.434 & \textbf{+0.067} \\
\addlinespace[2pt]
5\%   & SP        & 0.131 & 0.499 & +0.144 \\
5\%   & \pruningN & \textbf{0.050} & 0.315 & \textbf{+0.043} \\
\bottomrule
\end{tabular}
\caption{Matched sparsity: \pruningN\ and SP remove the same fraction of MLP
weights. At equal budget \pruningN\ reaches $2.6$--$4.0\times$ lower pre-attack ES
and $2.6$--$3.3\times$ lower $\Delta$ES. Its lower utility at the same budget
reflects a stronger intervention, which is why we also report the
matched-utility comparison in Table~\ref{tab:ctrl_utility}.}
\label{tab:ctrl_sparsity}
\end{table}

%% file: Tables/ctrl_matched_norm.tex
\begin{table}[H]
\centering
\small
\setlength{\tabcolsep}{5pt}
\begin{tabular}{lcccc}
\toprule
Edit & $\|\Delta W\|_2$ & pre-ES & pre-utility & $\Delta$ES \\
\midrule
SP 2.5\%        & 86.3  & 0.274 & 0.539 & +0.173 \\
\pruningN 1.5\% & 89.1  & \textbf{0.090} & 0.483 & \textbf{+0.079} \\
\addlinespace[2pt]
SP 5\%          & 120.1 & 0.131 & 0.499 & +0.144 \\
\pruningN 4\%   & 122.8 & \textbf{0.054} & 0.366 & \textbf{+0.049} \\
\bottomrule
\end{tabular}
\caption{Matched update norm: \pruningN\ and SP compared at near-identical
$\ell_2$ update magnitudes (gaps $\le 3\%$). \pruningN\ reaches $2.4$--$3.0\times$
lower pre-attack ES and $2.2$--$2.9\times$ lower $\Delta$ES, so displacement
magnitude alone does not explain the gain.}
\label{tab:ctrl_norm}
\end{table}

%% file: Tables/importance_variants.tex
\begin{table}[H]
\centering
\small
\setlength{\tabcolsep}{4pt}
\begin{tabular}{llc}
\toprule
Importance term & Requires & $\rho$ \\
\midrule
Fisher $g^2$                  & backward & $+0.57$ \\
Influence surrogate           & backward & $+0.25$ \\
Hessian $H_{jj}W^2$ (OBD)     & backward & $-0.79$ \\
Integrated grad.\ $|W\bar{g}|$   & backward & $-0.84$ \\
Activation $|W||X_f|$ (\predictorN) & forward & $\mathbf{-0.92}$ \\
\bottomrule
\end{tabular}
\caption{Replacing only \predictorN's importance term, evaluated on the same 15
TOFU-1B checkpoints and attack as Table~\ref{tab:rank_corr}. Negative is the
correct sign. Fisher and the influence surrogate get the sign wrong; the
Hessian and integrated-gradient variants are weaker and need backward passes.
The activation-based term is both the strongest and the only forward-only
choice.}
\label{tab:importance_variants}
\end{table}

%% file: Tables/lmc.tex
\begin{table}[H]
\centering
\small
\setlength{\tabcolsep}{4pt}
\begin{tabular}{lcc}
\toprule
Family & Checkpoints & $B_f$ \\
\midrule
Dense methods and \pruningN & 42 & $0.000$ \\
SP ($2.5$--$15\%$)               & 4  & $0.076$--$0.387$ \\
\bottomrule
\end{tabular}
\caption{Linear mode connectivity barrier $B_f$ between $\theta_0$ and the
unlearned model, ported from \citet{siddiqui2025dormant} to TOFU-1B. Dense methods are GradDiff, NPO and RMU. For every healthy non-SP checkpoint the loss curve shows no upward bump at all, so there
is nothing to rank by; only SP produces nonzero barriers, and there
\predictorN\ gives the same ordering.}
\label{tab:lmc}
\end{table}

%% file: Tables/muse_news.tex
\begin{table}[H]
\centering
\small
\setlength{\tabcolsep}{6pt}
\begin{tabular}{lccc}
\toprule
Method & Utility $\uparrow$ & pre-ES $\downarrow$ & $\Delta$ES $\downarrow$ \\
\midrule
GradDiff  & 0.048 & 0.008 & $+0.313$ \\
RMU       & 0.496 & 0.084 & $+0.205$ \\
SP        & 0.432 & 0.085 & $+0.072$ \\
\pruningN & 0.379 & \textbf{0.059} & $\mathbf{+0.067}$ \\
\bottomrule
\end{tabular}
\caption{Cross-benchmark check on MUSE-News (LLaMA-2-7B, retain-ROUGE utility,
gold $=0.557$; retain-only relearning attack, 1 epoch). GradDiff collapses in
utility; among the utility-preserving methods \pruningN\ has the lowest
$\Delta$ES.}
\label{tab:muse}
\end{table}